\documentclass{article}

\usepackage{spconf}
\usepackage[utf8]{inputenc}
\usepackage{amsmath,graphicx}

\usepackage{amsfonts} 
\usepackage{pifont}
\usepackage{bm}
\usepackage{comment}
\usepackage{url}

\renewcommand{\vec}[1]{\bm{#1}}

\newcommand{\mat}[1]{\bm{\mathrm{#1}}}

\newcommand{\T}{\mathsf{T}}

\title{Adaptive Dropout for Pruning Conformers}
\name{Yotaro Kubo, Xingyu Cai, Michiel Bacchiani}
\address{Google DeepMind \\
\texttt{\{yotaro, caixingyu, michiel\}@google.com}}

\begin{document}
\ninept
\maketitle
\begin{abstract}
This paper proposes a method to effectively perform joint \textit{training-and-pruning} based on adaptive dropout layers with unit-wise retention probabilities.
The proposed method is based on the estimation of a unit-wise retention probability in a dropout layer.
A unit that is estimated to have a small retention probability can be considered to be prunable.
The retention probability of the unit is estimated using back-propagation and the Gumbel-Softmax technique.
This pruning method is applied at several application points in Conformers such that the effective number of parameters can be significantly reduced.
Specifically, adaptive dropout layers are introduced in three locations in each Conformer block: (a) the hidden layer of the feed-forward-net component, (b) the query vectors and the value vectors of the self-attention component, and (c) the input vectors of the LConv component.
The proposed method is evaluated by conducting a speech recognition experiment on the LibriSpeech task.
It was shown that this approach could simultaneously achieve a parameter reduction and accuracy improvement.
The word error rates improved by approx 1\% while reducing the number of parameters by 54\%.
\end{abstract}

\section{Introduction}

Overparametrized models are important not only because of their expression ability but also for their desirable properties in gradient descent optimization \cite{du2018gradient}.
However, as a drawback, they are computationally expensive for both training and inference.
In speech applications, neural networks are often used on devices with limited computational capability, such as mobile phones and smart speakers \cite{he2019streaming}.
As a result, the accuracy benefits from overparametrized models can become impractical given the computational problem it introduces.

To prevent the large computational costs of overparametrized models, several methods have been proposed in the past, e.g., \cite{ma2023llm,sun2023simple}.
One of the most promising approaches is pruning the trained models.
Pruning is often effective because it is empirically known that hidden units of fine-tuned overparametrized models tend to be redundant.
Since pruning can still degrade the accuracy of the trained model, it's also common to perform post-training after pruning \cite{sun2023simple}. 
Alternatively, methods for performing pruning during training have also been investigated \cite{ma2019transformed}. The latter, joint training-and-pruning, is beneficial not only for accuracy, but also for simplicity of the overall training-deployment pipeline.

The granularity of the pruning is an algorithm parameter.
The finest granularity is at individual weights.
L$p$ regularization techniques ($p \leq 1$) can be used as a joint training-and-pruning method that prunes the individual weight parameters \cite{ma2019transformed,louizos2017learning}.
\cite{lecun1989optimal} identifies prunable weights by evaluating the second-order gradient with regards to each weight.
While the finest pruning can minimize the number of weight parameters, computational advantages of the weight-level pruned models can be limited.
This is because the weight-level sparsity can only be computationally efficient when using a special matrix multiplication algorithm.

Instead, we focus on a slightly more coarse-level and more hardware friendly pruning in this paper.
The number of hidden units, i.e. the dimensionality of internal representations, is minimized.
We refer to this as \textit{unit-level} pruning.
Unit-level pruning is important because typical deep learning accelerators, such as graphic-processing units (GPUs) and tensor-processing units (TPUs), do not support generic sparse matrix multiplication.

This paper extends the methods proposed in \cite{srinivas2016generalized,kubo2016compacting} and proposes a method to combine them within the latest speech encoder based on Conformers \cite{gulati2020conformer}.
Both \cite{srinivas2016generalized,kubo2016compacting} employ a dropout layer with unit-wise retention probabilities.
Such adaptive dropout layers (ADL) can make the units in the layers prunable.
If the retention probability of a unit converged to zero after training, the unit can be removed from the network without incurring an accuracy degradation.

The theoretical difference from the previous studies are the estimator and controllability of the algorithm.
The proposed method is based on the Gumbel-Softmax estimator \cite{jang2017categorical} that has been proven to be effective in many tasks (e.g. wav2vec-2.0 \cite{baevski2020wav2vec} and its applications).
The unit-wise retention probabilities are controlled to be close to the target value by L2 regularization.
Furthermore, this paper introduces a practical method to integrate ADLs in a Conformer block.
Previous studies showed the effectiveness of ADLs when integrated into a convolution layer \cite{srinivas2016generalized} and/or a fully-connected layer \cite{kubo2016compacting}.
This paper further applies it to multi-head self-attention (MHSA) and LConv layers in Conformer blocks.
With the application method proposed in the paper, almost all the parameters in a Conformer block are considered for pruning.

\section{Trainable adaptive dropout layers}

In this section, we first describe the formulation of ADL layers. We then describe the method to estimate the unit-wise retention probabilities in the ADL layers.

\subsection{Unit-wise dropout for pruning}

The dropout layer \cite{hinton2012improving} of neural networks can be expressed as follows:
\begin{equation}
\vec{y} = \vec{m} \odot \vec{x},
\end{equation}
where $\vec{x} \in \mathbb{R}^{D}$ and $\vec{y} \in \mathbb{R}^{D}$ are the input and output of the layer, respectively, and $\vec{m} \in \mathbb{R}^{D}$ is a vector of random samples from a Bernoulli distribution.

In unit-wise dropout layers, the $d$-th element of the mask vector $\vec{m}$ is drawn from the Bernoulli distribution with a parameter $\beta_d$ depending on $d$, as follows:
\begin{equation}
    m_d \sim \mathrm{Bern}(\varsigma(\beta_d)). \label{eq:bernoulli_mask}
\end{equation}

Here, $\varsigma$ is the sigmoid function, and $\mathrm{Bern}(\varsigma(\beta_d))$ denotes the Bernoulli distribution with the parameter $\varsigma(\beta_d)$.
A sample of $m_d$ has a value $1$ with the probability of $\varsigma(\beta_d)$, and $0$ with the probability of $1 - \varsigma(\beta_d)$.

The unit-wise dropout has different parameters $\beta_d$ for each $d$ unlike the conventional dropout.
The conventional dropout can be seen as a special case of the unit-wise dropout with $\beta_d = \beta_{d'}$ for all $d$ and $d'$.

The pruning can be performed by optimizing this $\beta_d$ with an additional regularization term for sparsity.
The unit-level sparsity can be achieved by encouraging $\beta_d$ to have a smaller value.
In a training session, if $\beta_d \to -\infty$, $\varsigma(\beta_d)$ converges to $0$.
This indicates that the $d$-th input to the dropout layer $x_d$ is no longer used in the following neural net components and hence can be removed.

\subsection{Back-propagation over retention rates}

In order to estimate the optimal $\beta_d$ with back-propagation, the Gumbel-Softmax technique (specifically, its binary variant, \textit{Gumbel-Sigmoid}) is applied.

The Gumbel-Sigmoid approach reparametrizes ${m}_d$ as follows:
\begin{equation}
    {m}_d = H({\beta}_d + \epsilon), \label{eq:reparametrized}
\end{equation}
where $H$ is a step function, i.e. $H(x) = 1$ for $x > 0$ and $H(x) = 0$ otherwise, and $\epsilon$ is a random variable following the standard logistic distribution with $\mu = 0$ and $s = 1$ (i.e. difference of two independent samples from the standard Gumbel distribution).
It is shown in \cite{jang2017categorical} that the distribution of $m_d$ follows the Bernoulli distribution defined in Eq.~\eqref{eq:bernoulli_mask}.

Because the step function $H$ in Eq.~\eqref{eq:reparametrized} is not differentiable, \cite{jang2017categorical} proposed to use a straight-through (ST) estimator.
The ST estimator enables back-propagation by approximating the non-differentiable part of the function by a differentiable counterpart when computing gradients.
The proposed method integrates the ST method by performing the exact computation of Eq.~\eqref{eq:reparametrized} for forward-propagation, and the following approximated computation in back-propagation:

\begin{equation}
    m_d = \varsigma(\beta_d + \epsilon), \label{eq:st_approximation}
\end{equation}
where $\varsigma$ is the sigmoid function.

In contrast to the method in \cite{srinivas2016generalized} which used a piece-wise linear function for computing the gradient, our method uses the sigmoid function to approximate the gradient computation.
As explained in the next section, it is especially important to avoid premature hard-decision in pruning.
Therefore, the sigmoid-ST approach is desirable because it can propagate gradients as long as the mask has a non-zero probability to be $1$.

\subsection{Controlled regularization of retention probability} \label{sec:regularization}

If no regularization is applied to $\beta_d$, the training process is expected to activate all hidden units because increasing the number of hidden units can directly lead to smaller losses.
Therefore, the training process is modified to prefer smaller $\beta_d$ values for our purposes.
This can be done by adding a regularization term to the loss function.

Both the previous studies \cite{srinivas2016generalized,kubo2016compacting} adopt the log-beta pdf as a fixed regularization term. While there is a theoretical advantage using a conjugate prior distribution, i.e. beta distribution, as a regularization term, there are some drawbacks in practice as well.

The first drawback is that it tends to ignore the warm-up period of the Conformer (and Transformer) training.
It is empirically known that a warm-up period is required for better training of Transformer-like neural nets.

If we are to do joint training-and-pruning, the desirable behavior is to adjust retention probabilities gradually during the convergence of the other parameters.
However, with a fixed regularization weight, the gradient of the regularization term becomes dominant in the early phase of training because the variance of the loss gradient is high.
Therefore, it's important to control the influence of the regularization term dynamically over the training process.

The second drawback is numerical instability as there are poles at 0 and 1 in the log-beta pdf with a typical hyper parameter setting \footnote{For sparsification, a beta distribution $\mathrm{Beta}(\alpha, \beta)$ with $\alpha < 1, \beta < 1$ is typically used and that has the poles.}.
Those poles make the retention rates converging quickly to the extremes at $0$ and $1$.
With a common gradient-based optimizer, this also causes premature hard-decisions of pruning rules.

Given those two drawbacks, the proposed method is designed based on an uncentered L2 regularization where the center point is controlled to prefer smaller values.
In particular, the following regularization term is added to the loss function:

\begin{equation}
R(\vec{\beta}; t) = \alpha \sum_{d} \left| \beta_d - c(t) \right|^2, \label{eq:uncentered_l2_regularization}
\end{equation}
where $\alpha$ is the regularization weight, $t$ is the number of steps in the optimization, and $c(t)$ is the target value at step $t$.

By designing $c(t)$ to be a decreasing function, the retention probability of the dropout is gradually decreasing during the optimization.
The convergence speed can be controlled by adjusting the speed of decrease of $c(t)$.
For example, a piece-wise linear scheduling function $c(t)$ can be defined as
\begin{equation}
    c(t) = \max\left\{ \frac{t}{K} c_{\infty} + \left(1 - \frac{t}{K}\right) c_{0}, c_{\infty}\right\}, \label{eq:piecewise_linear_scheduler}
\end{equation}
where $c_{0}, c_{\infty}, K$ are the hyperparameters of this scheduling function.
$c_{0}$ is an initial target bias, $c_{\infty} < c_0$ is a final target bias, and $K$ is a decreasing period.

It should be noted that the uncentered L2 regularization (Eq. \eqref{eq:uncentered_l2_regularization}) for $\beta_d$ can easily be implemented using a reparametrization technique that obtains $\beta_d$ from the raw parameter $\beta'_d$ as

\begin{equation}
    \beta_d = \sqrt{\frac{\gamma}{\alpha}}\beta'_d + c(t),
\end{equation}
where $\gamma$ is the L2 regularization weight for the other parameters.

The conventional L2 regularization term applied to $\beta'_d$ with the weight $\gamma$ can be rewritten as the regularization term in Eq. \eqref{eq:uncentered_l2_regularization}. 
Thus, no special implementation is required to apply the regularization to $\beta_d$.

Finally after training, one can remove the parameters of the hidden units that were disabled with high probability in training.
The decision can be made by introducing a threshold $\beta_{\mathrm{thres}}$ and examining $\beta_d < \beta_{\mathrm{thres}}$.
Unlike other methods that perform pruning and training in separate phases, fine-tuning is not necessary after pruning.

\section{Application to Conformer}

\begin{figure*}[tb]
    \centering
    \includegraphics[width=1.0\textwidth]{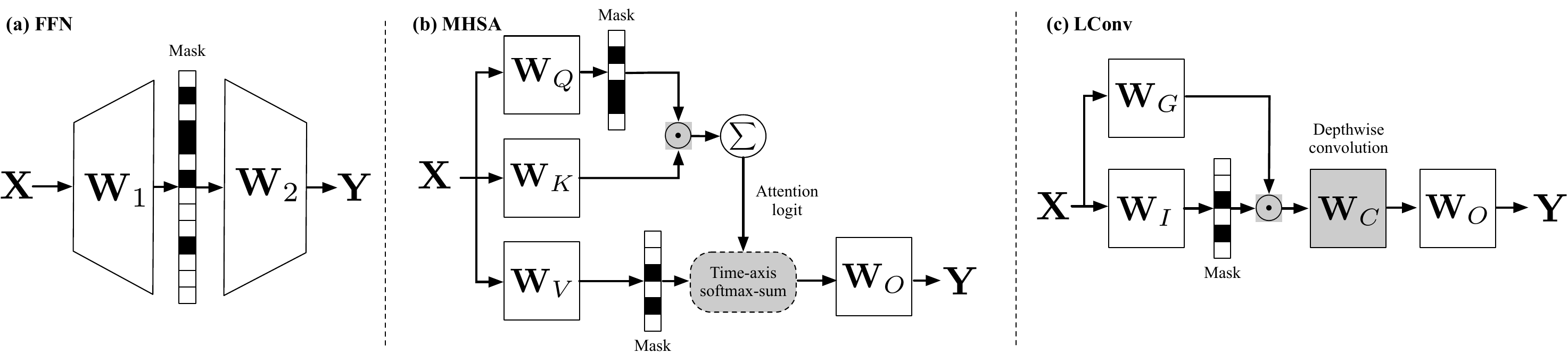}
    \caption{The locations of masking component inserted in each submodule of the Conformer. The parametrized components in a Conformer block are shown as boxes with the variable names in them (the bias parameters are omitted for simplicity.) The blocks with gray background represent dimensionality-wise computations. The effects of masking propagate beyond those gray blocks and make it possible to prune the parameters next to it.}
    \label{fig:diagram}
\end{figure*}

In this section, we propose a modification to Conformer blocks that integrates the ADLs described previously.
Except for the normalization layers, a Conformer block consists of the following sub-components:

\begin{itemize}
    \item two feed-forward networks (FFNs) each with a single hidden layer,
    \item an LConv module, and
    \item a multi-head self-attention (MHSA) module.
\end{itemize}

Fig.~\ref{fig:diagram} shows block diagrams of those sub-components with the ADL masks. 
The rest of this section explains how the parameters can be made prunable by introduction of the ADLs.

\paragraph*{FFN}
For FFNs, ADLs are inserted after the activation of the hidden layers.
Specifically, an FFN module with modification is enhanced as
\begin{equation}
    \vec{y} = \vec{b}_2 + \mat{W}_2 \left(\vec{m} \odot \vec{\phi}\left(\vec{b}_1 + \mat{W}_1 \vec{x}\right) \right),
\end{equation}
where $\vec{\phi}$ is an activation function, $\mat{W}_1, \mat{W}_2, \vec{b}_1$, and $\vec{b}_2$ are the parameters, and $\vec{m}$ is a mask vector drawn as Eq.~\eqref{eq:bernoulli_mask}. When $m_d$ is always $0$, $d$-th column of $\mat{W}_1$, $d$-th row of $\vec{b}_1$, and $d$-th row of $\mat{W}_2$ will not affect the output of this component, and therefore can be pruned.

\paragraph*{MHSA}
An MHSA block first extracts query and key vectors from the input.
The affine transforms for query and key vector extraction can be parametrized as $(\mat{W}_Q, \vec{b}_Q)$, and $(\mat{W}_K, \vec{b}_K)$ respectively \footnote{Some implementations of Transformers, including the one used in the experimental section \cite{paxml}, introduces additional scale factors when computing the inner-product between the query and key vectors. Those are often not prunable. For the sake of simplicity, those scale factors are omitted from the description.}.
The attention score $s^{(h)}_{t, t'}$ is then computed as
\begin{equation}
    s^{(h)}_{t, t'} =
    \left(\left( \vec{b}^{(h)}_{Q} + \mat{W}^{(h)}_Q\vec{x}_{t} \right)
    \odot \vec{m}^{(h)} \right)^{\T} 
    \left( \vec{b}^{(h)}_K + \mat{W}^{(h)}_K \vec{x}_{t'} \right),
\end{equation}
where the input $\vec{x}_{t}$ and $\vec{x}_{t'}$ are the $t$-th and $t'$-th frame in the input time-series $\mat{X}$, respectively, and $\vec{m}^{(h)}$ is the mask vector for $h$-th head drawn independently.

With the above modified formulation, if $m^{(h)}_d$ is always $0$, the $d$-th row of $\mat{W}^{(h)}_Q$, $\mat{W}^{(h)}_K$, $\vec{b}^{(h)}_Q$ and $\vec{b}^{(h)}_K$ can be pruned because those variables do not affect the attention score $s^{(h)}_{t, t'}$.

The attention probabilities $\bar{s}^{(h)}_{t, t'}$ are computed by applying scaled (and optionally time-masked) softmax to the above raw attention scores $s^{(h)}_{t, t'}$.
With the attention probabilities, the output of an MHSA block with the ADL can then be expressed as
\begin{equation}
\vec{y}_t = \vec{b}_O +  
\sum_{h} \mat{W}^{(h)}_O \sum_{t'}
\left( \bar{s}^{(h)}_{t, t'} \left(\vec{m}^{(h)} \odot \left( \vec{b}^{(h)}_V + \mat{W}^{(h)}_V \vec{x}_{t'} \right) \right) \right).
\end{equation}

Similar to the query and key vectors, when $m^{(h)}_d$ is always $0$, the $d$-th row of $\vec{b}^{(h)}_V$ and $\mat{W}^{(h)}_V$, and $d$-th column of $\mat{W}^{(h)}_O$ does not affect to the output and can be removed.

\paragraph*{LConv}

An LConv module can be decomposed into three parts: the input preparation part, depthwise 1d convolution part, and output mixing part.

The input preparation part computes the input to the convolution module using gated linear units (GLUs).
By masking the output of the GLUs using ADLs, the input preparation can be enhanced as
\begin{equation}
    \vec{z}_t = \left(\vec{b}_I + \mat{W}_I \vec{x}_t \right)
    \odot \left(\vec{b}_G + \mat{W}_G \vec{x}_t \right)
    \odot \vec{m}. \label{eq:lconv_first}
\end{equation}

Here, $\vec{z}_t$ is the input to the succeeding convolution module, $\mat{W}_I, \mat{W}_G, \vec{b}_I$, and $\vec{b}_G$ are the parameters of the input preparation, and $\vec{m}$ is the mask vector drawn from the ADL.

If $m_d = 0$, the $d$-th row of $\mat{W}_I$, $\mat{W}_G$, $\vec{b}_I$, and $\vec{b}_G$ do not affect the output $\vec{z}_t$, and therefore can be pruned.
Furthermore, $m_d = 0$ implies the $d$-th elements of $\vec{z}_t$s are zero for all $t$s.

A depthwise 1-d convolution without a bias term is applied as $\vec{z}'_{:, d} = \vec{w}_d *\vec{z}_{:, d}$ where $\vec{z}_{:, d}$ and $\vec{z}'_{:, d}$ are single channel time-series signals extracted from the $d$-th elements of the vectors in $\vec{z}_t$ and $\vec{z}'_t$, $\vec{w}_d$ is the convolution parameters, and $*$ is the 1-D convolution operator.
If $m_d$ in Eq. \eqref{eq:lconv_first} is zero, $\vec{z}_{:, d}$ is a constant zero, and therefore $\vec{z}'_{:, d}$ is also zero. Thus, the parameter $\vec{w}_d$ has no impact and can be removed.

The output of the LConv block can be expressed as
\begin{equation}
\vec{y}_t = \vec{b}_{O} + \mat{W}_{O} \vec{z}'_t.
\end{equation}

Since $m_d = 0$ implies that the $d$-th element of $\vec{z}'_t$ is zero, and the $d$-th column of $\mat{W}_O$ can be removed.

\begin{figure*}[tb]
    \centering
    \includegraphics[width=0.9\textwidth]{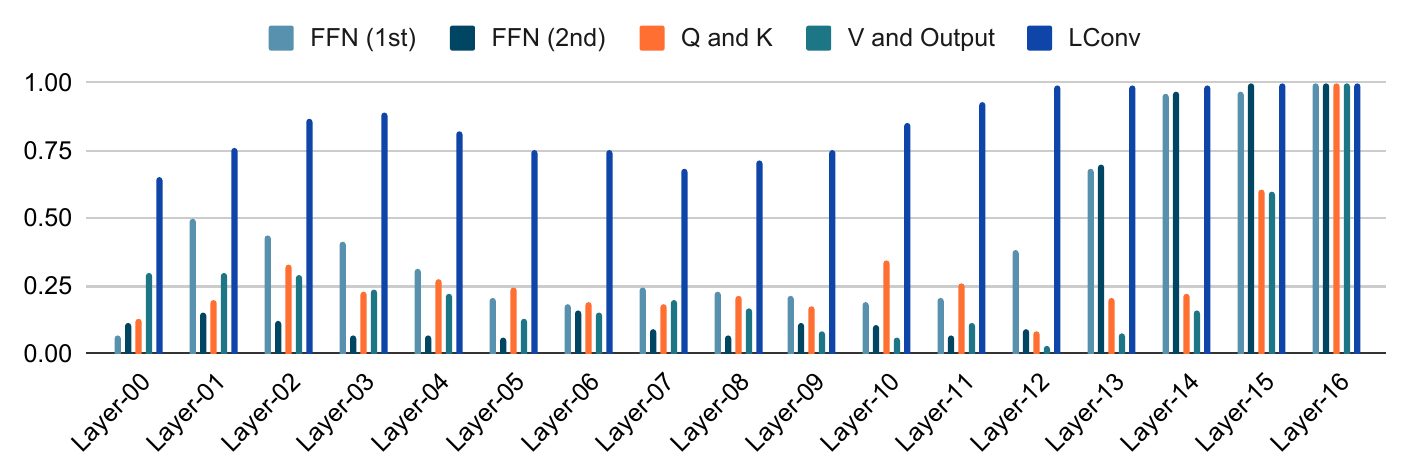}
    \caption{The ratio of surviving units for each different place where ADLs were inserted.}
    \label{fig:retention_rates}
\end{figure*}

\section{Experiments}

We evaluated the proposed method by pruning the {\tt ConformerL} architecture proposed in \cite{gulati2020conformer}.
The neural networks were trained to minimize a Connectionist Temporal Classification (CTC)-loss function \cite{graves2006connectionist} on the LibriSpeech \cite{panayotov2015librispeech} dataset.

We compared the proposed pruning method with the following hand-crafted compact models.
\begin{itemize}
    \item Shallow: {\tt ConformerL} with a smaller numbers of encoder blocks.
    \item Narrow: {\tt ConformerL} with a narrower model width. This also reduces the output size of the CNN feature processing module.
    \item NarrowFFN: {\tt ConformerL} with narrower FFNs. This variant uses 1024 hidden units in FFNs whereas the baseline {\tt ConformerL} uses 2048 units.
    \item NoMacaron: {\tt ConformerL} with a single FFN block for each block. Baseline Conformers use two FFNs that sandwiches MHSA and LConv blocks. In contrast, this variant only uses one after MHSA and LConv (similar to Transformers).
\end{itemize}

\begin{table}[bt]
    \centering
    \begin{tabular}{l|r|rrrr}
        \hline \hline
        & & \multicolumn{2}{c}{dev} & \multicolumn{2}{c}{test} \\
        & \#Params & clean & other & clean & other \\
        \hline
        ConformerL & 111M & 3.2 & 7.9 & 3.4 & 7.8 \\
        \hline
        Proposed & 50.1M & {\bf 3.0} & {\bf 7.7} & {\bf 3.2} & {\bf 7.7} \\
        \hline
        Shallow (L=9) & 62.7M & 3.2 & 8.5 & 3.5 & 8.4 \\
        Shallow (L=7) & 50.6M & 3.5 & 9.3 & 3.9 & 9.3 \\
        Narrow (D=384) & 62.7M & 3.2 & 8.2 & 3.4 & 8.4 \\
        Narrow (D=320) & 43.7M & 3.7 & 9.5 & 3.9 & 9.3 \\
        NarrowFFN & 75.5M & 3.6 & 9.1 & 3.9 & 9.3 \\
        NoMacaron & 75.5M & 3.4 & 9.0 & 3.8 & 8.9 \\
        \hline
    \end{tabular} 
    \caption{Word error rates of the proposed pruned model and hand-crafted small models.} \label{tab:all_wers}
\end{table}

We used the piece-wise linear scheduler (Eq. \eqref{eq:piecewise_linear_scheduler}) to
control the retention probabilities with the following hyper parameters: 
$c_0 = 10$, $c_{\infty} = -2$, and $K = 100000$.
The regularization weight in Eq. \eqref{eq:uncentered_l2_regularization} was fixed to $\alpha = 10^{-7}$ where the the L2 regularization coefficients for other variables were set to $\gamma = 10^{-5}$.

Table~\ref{tab:all_wers} summarizes the word error rates of the compared models.
The word error rates for "Proposed" is computed by pruning the units via setting $m_d = 0$ when $\beta_d < c_{\infty}$ and $m_d = 1$ otherwise.
The reported number of parameters for "Proposed" is the number of effective parameters, i.e., total parameters minus the parameters associated with $\beta_d < c_{\infty}$.

From the table, we observed that the proposed model achieved the best results among the compared models, even when compared to the original dense model.
This improvement can be attributed to the regularization effect of the dropout components.
Even though the dropout layers were also used in the baseline mode, only the fixed layer-wise retention probabilities were considered in the baseline.
As suggested in \cite{rennie2014annealed}, dropout regularization can especially be important in the early phase of training.
Our method applies dropout in the early phase of training as well, and gradually reduces the degree of uncertainty as optimization progresses.
We believe that this is why our model was better than the baseline model with the hand-tuned retention probabilities.

Furthermore, we confirmed that larger models are beneficial for improving generalization.
The narrower models, and shallower models we tested in the experiments did not perform better compared to the original model.
Thus, we can conclude that the proposed method could effectively utilize the advantages of overparametrized models yet could yield a compact model.

The ratio of surviving units that were not pruned after training is summarized in Fig.~\ref{fig:retention_rates}.
By comparing the survival rates in the different components, it was observed that units in the FFN and MHSA components tended to be pruned more than in the other components.
The lower rates in the FFN components can be attributed to the fact that the number of FFNs in Conformers is twice as large than in Transformers.
Therefore, FFNs in the Conformers are highly redundant by design.
However, hand-optimizing the topology by just removing the first FFN (similar to the Transformer architecture) was found to be less effective ("NoMacaron" in Table~\ref{tab:all_wers}).
This result can be interpreted as additional evidence supporting the effectiveness of the strategy that starts training from the redundant overparametrized model and then applies pruning.

Furthermore, as shown in the figure, the ratio of the surviving units is considerably smaller in the input-side layers of the Conformer.
Especially, it was observed that the first FFN of the first layer was almost completely disabled  (the unit survival rate was only 7\%).
This suggest that a component-level pruning technique similar to neural architecture search \cite{wistuba2019survey} can also be achieved using ADLs.
Developing an architecture search using ADLs is left as a future work.

\section{Conclusions}

In this paper, we proposed a novel algorithm to prune hidden units in neural networks, and applied it to compact Conformer-based speech encoders.
The proposed method is based on an enhanced dropout technique with unit-wise retention probabilities.
A method for effectively estimating a sparse unit-wise retention probability via back-propagation is proposed.
In order to improve the compatibility with the warm-up process of the Conformer (and Transformer) training, a scheduling function that gradually emphasizes a sparsity constraint is introduced in the training process.
The experimental results showed that the proposed method can compress the best performing Conformer encoder by half with regard to the number of parameters and simultaneously achieve better accuracies.

\bibliographystyle{IEEEbib}
\bibliography{refs}

\begin{thebibliography}{10}

\bibitem{du2018gradient}
Simon~S. Du, Xiyu Zhai, Barnab{\'{a}}s P{\'{o}}czos, and Aarti Singh,
\newblock ``Gradient descent provably optimizes over-parameterized neural
  networks,''
\newblock in {\em Proc. ICLR}, 2018.

\bibitem{he2019streaming}
Yanzhang He, Tara~N. Sainath, Rohit Prabhavalkar, Ian McGraw, Raziel Alvarez,
  Ding Zhao, David Rybach, Anjuli Kannan, Yonghui Wu, Ruoming Pang, Qiao Liang,
  Deepti Bhatia, Yuan Shangguan, Bo~Li, Golan Pundak, Khe~Chai Sim, Tom Bagby,
  Shuo{-}Yiin Chang, Kanishka Rao, and Alexander Gruenstein,
\newblock ``Streaming end-to-end speech recognition for mobile devices,''
\newblock in {\em Proc. ICASSP}, 2019.

\bibitem{ma2023llm}
Xinyin Ma, Gongfan Fang, and Xinchao Wang,
\newblock ``{LLM-Pruner}: On the structural pruning of large language models,''
\newblock in {\em Advances in neural information processing systems}, 2023.

\bibitem{sun2023simple}
Mingjie Sun, Zhuang Liu, Anna Bair, and J~Zico Kolter,
\newblock ``A simple and effective pruning approach for large language
  models,''
\newblock in {\em Proc. ICLR}, 2024.

\bibitem{ma2019transformed}
Rongrong Ma, Jianyu Miao, Lingfeng Niu, and Peng Zhang,
\newblock ``Transformed $\ell$1 regularization for learning sparse deep neural
  networks,''
\newblock {\em Neural Networks}, vol. 119, pp. 286--298, 2019.

\bibitem{louizos2017learning}
Christos Louizos, Max Welling, and Diederik~P. Kingma,
\newblock ``Learning sparse neural networks through $ l\_0 $ regularization,''
\newblock {\em arXiv preprint arXiv:1712.01312}, 2017.

\bibitem{lecun1989optimal}
Yann LeCun, John~S. Denker, and Sara~A. Solla,
\newblock ``Optimal brain damage,''
\newblock {\em Advances in neural information processing systems}, 1989.

\bibitem{srinivas2016generalized}
Suraj Srinivas and R.~Venkatesh Babu,
\newblock ``Generalized dropout,''
\newblock {\em arXiv preprint arXiv:1611.06791}, 2016.

\bibitem{kubo2016compacting}
Yotaro Kubo, George Tucker, and Simon Wiesler,
\newblock ``Compacting neural network classifiers via dropout training,''
\newblock {\em arXiv preprint arXiv:1611.06148}, 2016.

\bibitem{gulati2020conformer}
Anmol Gulati, James Qin, Chung{-}Cheng Chiu, Niki Parmar, Yu~Zhang, Jiahui Yu,
  Wei Han, Shibo Wang, Zhengdong Zhang, Yonghui Wu, and Ruoming Pang,
\newblock ``Conformer: Convolution-augmented transformer for speech
  recognition,''
\newblock in {\em Proc. INTERSPEECH}, 2020, pp. 5036--5040.

\bibitem{jang2017categorical}
Eric Jang, Shixiang Gu, and Ben Poole,
\newblock ``Categorical reparameterization with gumbel-softmax,''
\newblock in {\em Proc. ICLR}, 2017.

\bibitem{baevski2020wav2vec}
Alexei Baevski, Yuhao Zhou, Abdelrahman Mohamed, and Michael Auli,
\newblock ``wav2vec 2.0: {A} framework for self-supervised learning of speech
  representations,''
\newblock in {\em Advances in neural information processing systems}, 2020.

\bibitem{hinton2012improving}
Geoffrey~E. Hinton, Nitish Srivastava, Alex Krizhevsky, Ilya Sutskever, and
  Ruslan Salakhutdinov,
\newblock ``Improving neural networks by preventing co-adaptation of feature
  detectors,''
\newblock {\em arXiv preprint arXiv:1207.0580}, 2012.

\bibitem{paxml}
Google,
\newblock ``{PaxML},'' \url{https://github.com/google/paxml}.

\bibitem{graves2006connectionist}
Alex Graves, Santiago Fern{\'{a}}ndez, Faustino~J. Gomez, and J{\"{u}}rgen
  Schmidhuber,
\newblock ``Connectionist temporal classification: labelling unsegmented
  sequence data with recurrent neural networks,''
\newblock in {\em Proc. ICML}, 2006.

\bibitem{panayotov2015librispeech}
Vassil Panayotov, Guoguo Chen, Daniel Povey, and Sanjeev Khudanpur,
\newblock ``Librispeech: An {ASR} corpus based on public domain audio books,''
\newblock in {\em Proc. ICASSP}, 2015.

\bibitem{rennie2014annealed}
Steven~J. Rennie, Vaibhava Goel, and Samuel Thomas,
\newblock ``Annealed dropout training of deep networks,''
\newblock in {\em Proc. IEEE Spoken Language Technology Workshop (SLT)}, 2014.

\bibitem{wistuba2019survey}
Martin Wistuba, Ambrish Rawat, and Tejaswini Pedapati,
\newblock ``A survey on neural architecture search,''
\newblock {\em arXiv preprint arXiv:1905.01392}, 2019.

\end{thebibliography}
\end{document}